\documentclass[sigconf]{acmart}
\def\arxivversion{1}
\def\longversion{1}

\if\arxivversion1
    \renewcommand\footnotetextcopyrightpermission[1]{}
\fi

\usepackage[normalem]{ulem}
\usepackage{multirow}
\usepackage{multicol}
\usepackage{siunitx}
\usepackage{bm}

\usepackage{tikz}
\usetikzlibrary{
    calc,
    decorations.markings,
    math,
    matrix,
    chains,
    positioning,
    arrows,
    shapes,
    decorations.pathreplacing,
    arrows.meta,
    shapes.multipart
}

\usepackage{listings}
\definecolor{kwordargument}{RGB}{102,0,153}
\definecolor{strings}{RGB}{0,128,128}
\definecolor{keyword}{RGB}{0,0,128}
\lstdefinestyle{mystyle}{
  keywordstyle=\color{keyword},
  stringstyle=\color{strings},
  basicstyle=\ttfamily\footnotesize,
  breakatwhitespace=false,         
  breaklines=true,                 
  captionpos=b,                    
  keepspaces=true,                 
  numbersep=5pt,                  
  showspaces=false,                
  showstringspaces=false,
  showtabs=false,                  
  tabsize=2,
  emph={predictions,frames},          
  emphstyle=\color{kwordargument},    
}
\newrobustcmd*{\bftabnum}{%
	\bfseries
	\sisetup{output-decimal-marker={\textbf{.}}}%
}

\AtBeginDocument{%
  \providecommand\BibTeX{{%
    \normalfont B\kern-0.5em{\scshape i\kern-0.25em b}\kern-0.8em\TeX}}}

\setcopyright{acmcopyright}
\copyrightyear{2020}
\acmYear{2020}
\acmDOI{XX.XXXX/XXXXXXX.XXXXXXX} 

\acmConference[ACM MM '20]{ACM MM '20: ACM Multimedia 2020}{October 12--16, 2020}{Seattle, United States}
\acmBooktitle{ACM MM '20: ACM Multimedia 2020, October 12--16, 2020, Seattle, United States}
\acmPrice{15.00}
\acmISBN{XXX-X-XXXX-XXXX-X/20/XX}

\begin{document}

\title{TransNet V2: An effective deep network architecture for fast shot transition detection}


\author{Tom\'{a}\v{s} Sou\v{c}ek and Jakub Loko\v{c}}
\email{tomas.soucek1@gmail.com, lokoc@ksi.mff.cuni.cz}
\affiliation{%
  \institution{SIRET Research Group, Department of Software Engineering \\
Faculty of Mathematics and Physics, Charles University, Prague, Czech Republic}
}


\begin{abstract}
Although automatic shot transition detection approaches are already investigated for more than two decades, an effective universal human-level model was not proposed yet. Even for common shot transitions like hard cuts or simple gradual changes, the potential diversity of analyzed video contents may still lead to both false hits and false dismissals. Recently, deep learning-based approaches significantly improved the accuracy of shot transition detection using 3D convolutional architectures and artificially created training data. Nevertheless, one hundred percent accuracy is still an unreachable ideal. In this paper, we share the current version of our deep network TransNet V2 that reaches state-of-the-art performance on respected benchmarks. A trained instance of the model is provided so it can be instantly utilized by the community for a highly efficient analysis of large video archives. Furthermore, the network architecture, as well as our experience with the training process, are detailed, including simple code snippets for convenient usage of the proposed model and visualization of results.
\end{abstract}



\keywords{video processing, neural networks, shot boundary detection}


\maketitle

\section{Introduction}

Automatic identification of shot transitions in a video is a classical task of video analysis, where various traditional (even simple) algorithms provided results deemed as sufficient approximations at some popular benchmarks \cite{SMEATON2010411}. However, the benchmarks do not cover all ``open-world'' aspects of video content, and so the problem cannot be considered as solved yet. For example, fast abrupt changes of the visual contents caused by camera/environment/entity can still confuse even state-of-the-art shot transition detection models and may lead to many false hits. Besides false hits, trained models can sometimes miss a difficult (e.g. long) transition, even if such type was present in the utilized train set. In addition, novel ``wild'' transition types supported in common video editing tools and often used in television content may find shot transition detection models ``unprepared'' for a given type. Hence, it is necessary to search for more advanced approaches that could push the accuracy of shot transition detectors to new levels.

Currently, end-to-end deep learning approaches have become a mainstream research direction for video analysis tasks. For transition detection, temporal context is essential, and so a detection approach requires either an aggregation of extracted features from individual frames \cite{Karpathy_2014_CVPR} or utilization of 3D convolutions that jointly process spatial and temporal information.
The latter approach, popularized by Tran et al. \cite{tran2015learning} for video classification, was also considered for shot boundary detection networks proposed by Hassanien et al. \cite{HassanienESHM17} and Gygli \cite{Gygli17}.
Hassanien et al. predict a likelihood of sharp or gradual transition in a 16 frame sequence by the C3D network \cite{tran2015learning}. The predictions are, however, not used directly, and an SVM classifier is trained to give a labeling estimate. Further, some false positives are suppressed by color histogram differencing. Gygli, on the other hand, utilizes only predictions from a 3D convolutional network without any post-processing. However the network is much smaller and outperformed by Hassanien et al. Our previously proposed shot transition detection 3D convolutional network TransNet \cite{transnet} combines the end-to-end no-post-processing approach of Gygli with performance comparable to Hassanien et al. on RAI dataset \cite{Baraldi15RAI}.

In this paper, we present a new, improved version of TransNet architecture. The new model, labeled as ``TransNet V2'', is provided as an open-source project with an easy-to-use trained instance and utilized training/evaluation codes (all available at \url{https://github.com/soCzech/TransNetV2}). The current version provides promising detection accuracy and enables efficient processing of larger datasets\footnote{For example, we employ TransNet to analyze the V3C1 collection \cite{V3C}, which is currently used at the Video Browser Showdown \cite{9037125}}. Given three different benchmark datasets (ClipShots \cite{Tang2018clipshots}, BBC \cite{Baraldi2015SceneSiamDet_BBC}, RAI \cite{Baraldi15RAI}), we also demonstrate that the new TransNet version represents a state-of-the-art approach. We  emphasize that all compared related models were re-evaluated (given available info) on the benchmarks, and the results were processed with the same evaluation script.

\section{TransNet V2}
In this section, we present our enhanced model TransNet V2 for shot detection, summarize key changes made to the architecture, and detail the training process. We also re-evaluate selected related work architectures and present the results at the end of the section.
More details on the architecture and comprehensive evaluations are included in the forthcoming thesis of T. Sou\v{c}ek \cite{soucek2020}.

\subsection{Architecture}
\subsubsection{Overview}
The proposed TransNet V2 builds on basic original TransNet concepts \cite{transnet}, where a resized input sequence of frames is initially processed with \emph{Dilated DCNN} cells. Specifically, the previously released TransNet version comprises six DDCNN cells where every cell consists of four $3 \times 3 \times 3$ convolution operations, each with $F$ filters and different dilation rates $1,\ 2,\ 4,\ 8$ for the temporal dimension. Hence, a larger receptive field of 97 frames is reached by the final sixth TransNet's DDCNN cell while using still an acceptable number of learnable parameters. In the new version, DDCNN cells also incorporate batch normalization that stabilizes gradients and adds noise during training. Every second cell contains a skip connection followed by spatial average pooling that reduces spatial dimension by two, as illustrated, with additional improvements, in the overall schema of the shared TransNet V2 network instance in Figure~\ref{fig:transnetv2_architecture}. Due to the lack of space, only the key additional concept changes to the architecture are further detailed in the following paragraphs.

\subsubsection{Convolution Kernel Factorization}
Xie et al. \cite{Xie_2018_ECCV} show that it might be beneficial to disentangle 3D $k \times k \times k$ convolutions into a 2D $k\times k$ spatial convolution followed by a 1D temporal convolution with kernel size $k$. Such disentanglement of the 3D convolutional kernel forces separate learning of image feature extraction and temporal comparison of the inferred features. Furthermore, if a low-enough number of spatial convolution filters is used, factorized convolutions reduce the number of learnable parameters, which may prevent over-fitting to artificially generated training data.

\subsubsection{Frame Similarities as Features}
\begin{figure}
    \centering
    \includegraphics[width=0.47\textwidth]{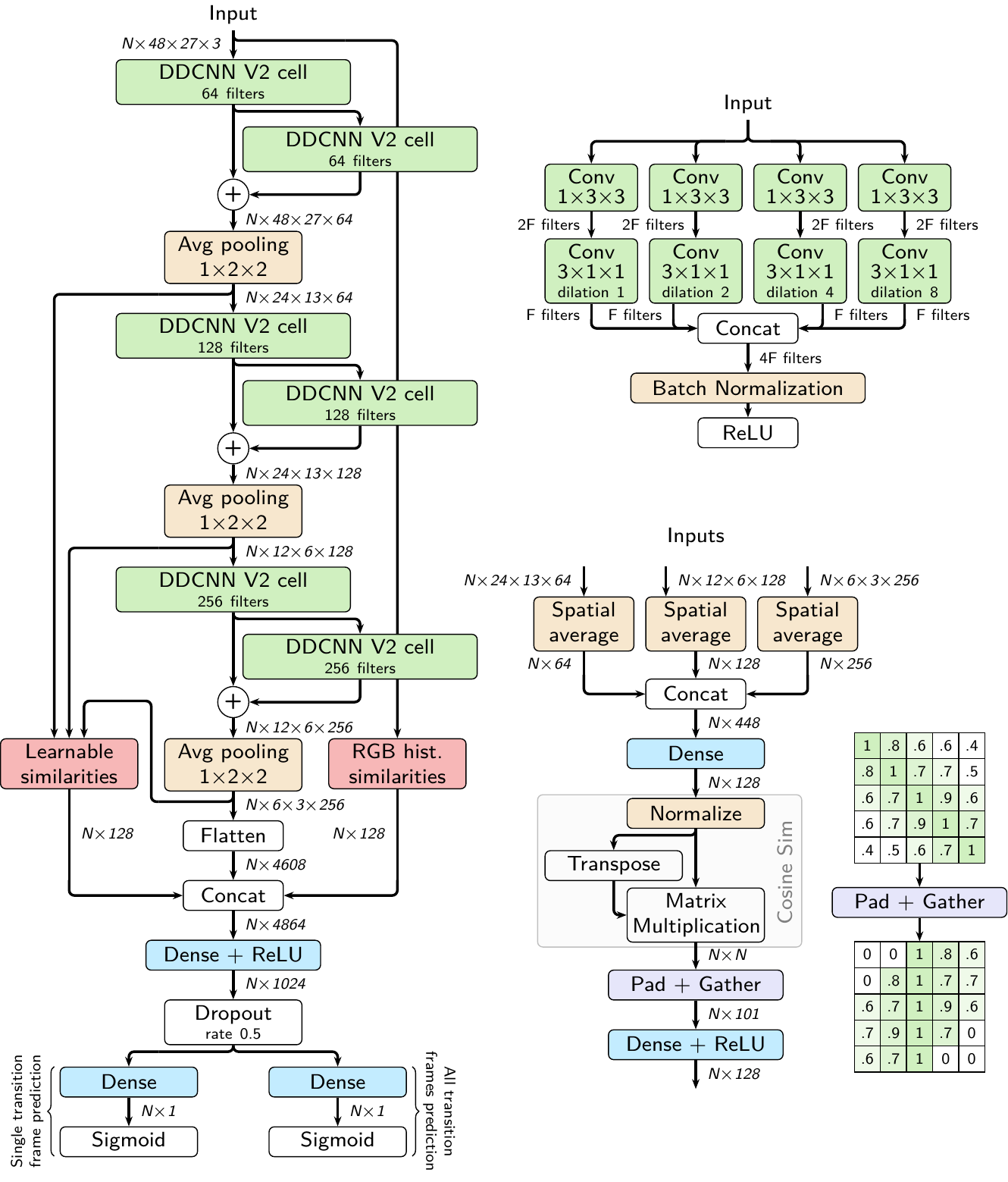}
    \caption{TransNet V2 Architecture (left), DDCNN V2 cell (right top), and learnable frame similarities computation (right bottom) with visualization of Pad + Gather operation.}
    \label{fig:transnetv2_architecture}
\end{figure}

Handcrafted and/or learnable features provide another option to design a shot transition detector relying on similarity scores between consecutive frames.
Therefore the new version of TransNet also incorporates RGB color histograms ($8^3=512$ bins) as well as learned features computed by spatially averaging activations of each average pooling and projecting them by a single dense layer. The features are processed by a similarity evaluation network node (denoted with red color in Figure~\ref{fig:transnetv2_architecture}), where the cosine similarity matrix is evaluated for collected features of processed frames. Each frame is then represented by similarity to its 50 preceding and following frames\footnote{Unavailable similarity values are replaced with zeros.}. The similarity vector is further transformed with a dense layer and concatenated to other inferred features from other parts of the network. Let us note that using similarity vectors was already proposed by a traditional shot transition detection approach \cite{CHASANIS_SVMonSimilarityVectors}.

\subsubsection{Multiple Classification Heads} 
TransNet V2 relies on two prediction heads. One head is trained to predict only a single middle frame of a transition, no matter its length. The second head predicts all transition frames. However, the only purpose of the second head is to update network weights during training to improve the network's ``understanding'' of what constitutes a transition and how long the transition is. More details about the prediction process are presented in Section~\ref{sec:reeval_protocol}.

\subsection{Training Setup and Experience}
\subsubsection{Train datasets.}
To train TransNet V2 classification heads, it is essential to acquire a large annotated train dataset comprising both hard cuts and gradual transitions. We follow the trend of synthetic training transitions \cite{HassanienESHM17,Gygli17} that can be rendered on the fly for pairs of randomly selected shots from a large reference shot collection. Specifically, the TRECVID IACC.3 \cite{2017trecvidawad} is used as it provides the master shot reference with several hundred thousand shots. We also experimented with a larger dataset of real transitions extracted from the ClipShots dataset \cite{Tang2018clipshots} with 4039 videos, where 128636 hard cuts and 38120 gradual transitions are manually annotated. However, according to our evaluations presented in Table \ref{tb:train_data}, synthetically rendered transitions significantly boost performance, and so the training process uses just 15\% of real ClipShots transitions compared to 85\% of synthetic transitions (35\% hard cuts, 50\% dissolves) generated from both IACC.3 and ClipShots. In addition, a significant proportion of gradual dissolve transitions in the train set seems to be important as well. Following our experience with TransNet, all videos are resized to the lower resolution 48x27.

\begin{table}[t]
	\centering
	{\small
	\begin{tabular}{l@{\hspace{0.3cm}}cccc}
		\toprule
		\textbf{Training data} & ClipShots & BBC  & RAI \\
		\midrule
		100\% real transitions                       & $66.4 \pm 1.3$ & $\bm{96.3} \pm 0.5$ & $86.4 \pm 1.2$ \\
		50\% real, 50\% cuts                         & $68.2 \pm 0.7$ & $\bm{96.6} \pm 0.7$ & $84.4 \pm 0.6$ \\
		50\% real, 50\% dissolves                    & $75.3 \pm 0.8$ & $\bm{96.3} \pm 0.5$ & $90.7 \pm 0.7$ \\
		15\% real, 35\% cuts, 50\% dissolves         & $77.0 \pm 0.8$ & $\bm{96.5} \pm 0.5$ & $91.2 \pm 1.1$ \\
		50\% cuts, 50\% dissolves                    & $\bm{77.5} \pm 0.3$ & $95.1 \pm 0.5$ & $\bm{93.2} \pm 0.9$ \\
		
		\bottomrule
	\end{tabular}
	}
	\caption{Effects of real and synthetic transitions. Mean F1 scores and standard deviations computed from 3 best epochs of 3 independent runs as measured on validation set.}
	\label{tb:train_data}
\end{table}

\subsubsection{Train sequence rendering.}
For real transitions, a sequence of 100 frames containing an annotated transition is randomly selected by the training script.
For the synthetic train set, we extract 300 frame segments from each reference dataset scene. The frames are extracted from the start, center, and end part of the scene while omitting some if the scene is not long enough. During training, two random segments are selected and randomly cropped to the length of 100 frames. The two segments are joined by a random transition type at a random position. Specifically, hard cuts and dissolves spanning over 2 to 30 frames are generated. Sequences without any transition are not employed for training as we assume that hard negatives are contained in already used input sequences.

\subsubsection{Data augmentation.}
We apply standard image augmentation to all frames in a shot. In order to prevent an introduction of fake shot boundaries into an input frame sequence, every image in the sequence is augmented in the same way. Firstly, shot frames are flipped left to right with probability $0.5$ and top to bottom with probability $0.1$. Further, standard TensorFlow image operations adjusting saturation, contrast, brightness, and hue are applied. With probability $0.05$ we also apply \textit{Equalize}, \textit{Posterize} and \textit{Color} operations from Python image library PIL\footnote{\url{https://pillow.readthedocs.io}, re-implemented in TensorFlow at \url{https://github.com/tensorflow/tpu/blob/master/models/official/efficientnet/autoaugment.py}.}. In the case of automatic transition generation, a color transfer technique is utilized on 10\% of input sequences before shot joining. The color transfer simulates a similar appearance of shots in one video.

\subsubsection{Technical details.}
We train both classification heads using standard cross-entropy loss averaged over batch. The positive class in the first \textit{single-frame} head is weighted by a factor of $5$. The second \textit{all-frame} head's contribution to the loss is discounted by $0.1$. L2 regularization is added to the loss weighted by $0.0001$. We optimize the loss function by SGD with momentum set to $0.9$ and a fixed learning rate of $0.01$. We train the network for 50 epochs, each with 750 batches of size 16 (in total 600,000 transitions). The best performing model on our ClipShots validation set (a subset of the official train set) is selected. Together with validation, the training takes approximately 17 hours on a single Tesla V100 16GB GPU. TensorFlow deep learning library was used for all the experiments.

\subsection{Comparison To Related Work}
To relate the effectiveness of TransNet models to a baseline, we identified two state-of-the-art models, mostly according to reported F1 scores on the popular RAI dataset. Specifically, DeepSBD by Hassanien et al. \cite{HassanienESHM17} reporting F1 score 93.4\% and DSM by Tang et al. \cite{Tang2018clipshots} reporting F1 score 93.5\%. However, for two additional ClipShots and BBC datasets, the F1 scores are not reported by Hassanien et al., while the evaluation code of Tang et al.\footnote{\label{note:1}Available at \url{https://github.com/Tangshitao/ClipShots_basline}.} is inconsistent with our evaluation method\footnote{We use the same evaluation metric as Baraldi et al. \cite{Baraldi15RAI} and the original TransNet~\cite{transnet}. However, due to minor errors in ground truth of some test sets, we also count correctly any detection that misses ground truth by at most two frames. With correct ground truth its effect compared to the original metric is minimal.\label{fn:metric}} following Baraldi et al. \cite{Baraldi15RAI}.
Therefore, we decided to re-evaluate both methods from available shared repositories and report results with a unified scoring methodology. We  emphasize that not all information was available and so the presented results do not have to correspond to the most optimal setting of related methods.

\subsubsection{Re-evaluation Protocol for Related Models}\label{sec:reeval_protocol}
Although the description of DeepSBD includes SVM and post-processing steps, the publicly available code\footnote{Available at \url{https://github.com/melgharib/DSBD}.} contains just the network structure. Hence, we utilize only its softmax predictions.
Since the model distinguishes between cut and gradual transition types, we sum their predicted confidence scores into a single ``transition class''.
To generate per-frame predictions, we assign the predicted confidence score to the middle 8 frames of the 16 frame input sequence (the input video sequence is always shifted by 8 frames). The final prediction is then thresholded, and the F1 score is computed for all test sets. Multiple thresholds were investigated, and the value of $0.9$ was chosen as it achieved the best overall results on the test sets. Therefore, the resulting model's performance can be overestimated. On the other hand, it raises the bar that could be potentially surpassed by a competitor. Please also note this evaluation approach even slightly surpasses the reported performance of DeepSBD on the RAI dataset (93.4\% vs. 93.9\%). Hence, we are more confident that our re-evaluation may objectively correspond to the work of Hassanien et al.

Similar difficulties appeared also for the DSM model \cite{Tang2018clipshots}, where the implementation of the multi-step process for shot boundary detection was not available as well. The authors pointed us to their ResNet-18 baseline, being the only publicly available code\textsuperscript{\ref{note:1}}. Regarding evaluations, the same approach, as in the case of DeepSBM, was taken. The best performing threshold of $0.8$ was again determined on the test sets.

TransNet V2 follows the same evaluation procedure as the original TransNet. The confidence of a transition appearance is predicted for all $N=100$ input frames; however, only the middle 50 predictions are considered for testing due to limited temporal context for the remaining predictions. While the original TransNet uses threshold $0.1$, TransNet V2 uses a fixed threshold of $0.5$. Only the \textit{single-frame} head confidence scores are utilized for transition prediction even though information from the second head could potentially improve the model's performance.

\subsubsection{Shot detection and results.}
Given confidence scores predicted by a network for every video frame and a threshold $\theta$, all the longest consecutive frame sequences with confidence scores greater than $\theta$ are declared as transitions. A shot is formed by all frames between two detected transitions.
In Table \ref{tb:results}, TransNet V2 is compared to selected related models on the official ClipShots test set \cite{Tang2018clipshots}, BBC Planet Earth documentary series \cite{Baraldi2015SceneSiamDet_BBC}, and RAI dataset \cite{Baraldi15RAI}. Our evaluations show that TransNet V2 is preferable for both ClipShots and BBC Planet Earth datasets and comparable to other models on the RAI dataset. All compared methods have lower performance on ClipShots, which is the largest collection, and so manual validation of the ground truth might be a challenge. According to our observations, some video parts remain unannotated, and some frames are incorrectly labeled as transition frames.

\begin{table}
	\centering
	\sisetup{detect-weight=true,detect-inline-weight=math}
	\begin{tabular}{l@{\hspace{0.3cm}}S[table-format=2.1]S[table-format=2.1]S[table-format=2.1]}
		\toprule
		\textbf{Model} & \multicolumn{1}{c}{ClipShots} & \multicolumn{1}{c}{BBC}  & \multicolumn{1}{c}{RAI} \\
		\midrule
		TransNet \cite{transnet}                     & 73.5 & 92.9 &  \bftabnum 94.3 \\
        Hassanien et al. \cite{HassanienESHM17}             & 75.9\textsuperscript{$*$} & 92.6\textsuperscript{$*$} & 93.9\textsuperscript{$*$} \\
        Tang et al. \cite{Tang2018clipshots}, ResNet baseline & 76.1\textsuperscript{$*$} & 89.3\textsuperscript{$*$} & 92.8\textsuperscript{$*$} \\
        \textbf{Ours} (TransNet V2)          & \bftabnum 77.9 & \bftabnum 96.2 & 93.9 \\
		\bottomrule
		\multicolumn{4}{l}{\footnotesize \textsuperscript{$*$} Our reevaluation. The best threshold selected on the test sets.}  \\
	\end{tabular}
	\caption{TransNet V2 compared to related works (F1 scores\textsuperscript{\ref{fn:metric}}).}
	\label{tb:results}
\end{table}

\subsection{Simple Usage Interface}
\begin{figure}
    \centering
    \if\longversion1
        \includegraphics[width=0.47\textwidth]{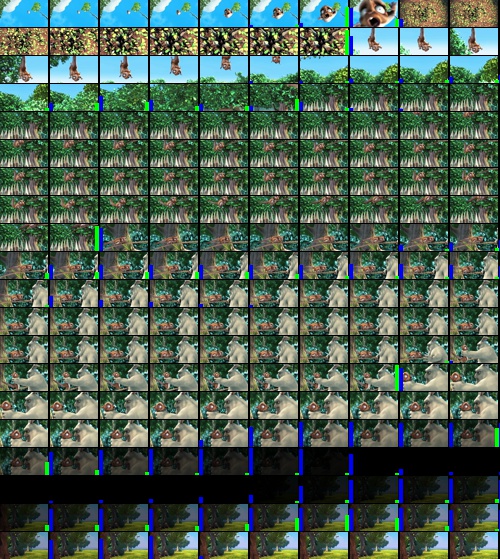}
        List of scenes (indexed from zero, both limits inclusive):
        $$(  0,  6)\ (  7,  16)\ ( 17,  81)\ ( 82, 137)\ (138, 159)\ (160, 186)\ (187, 199)$$
    \else
        \includegraphics[width=0.47\textwidth]{pdf/video_compressed.jpg}
        List of scenes (indexed from zero, both limits inclusive):
        $$(  0,  6)\ (  7,  16)\ ( 17,  61)\ ( 62, 97)\ (98, 119)\ (120, 146)\ (147, 159)$$
    \fi
    \caption{Visualized predictions from both classification heads with a corresponding list of scenes. The original video authored by Blender Foundation licensed under CC-BY. Sequences with no transitions shortened due to limited space.}
    \label{fig:example_predictions}
\end{figure}

We provide both the full training and evaluation code available at \url{https://github.com/soCzech/TransNetV2}. We also provide the trained TransNet V2 weights and especially simple API in \texttt{inference} directory of the repository. Detecting shots in a video is as follows:

\begin{lstlisting}[language=Python]
    from transnetv2 import TransNetV2
    model = TransNetV2("/path/to/weights_dir/")
    video_frames, single_frame_p, all_frame_p = \
        model.predict_video("/path/to/video.mp4")
\end{lstlisting}
We show the visualization of the model's predictions together with its list of scenes in Figure \ref{fig:example_predictions}, both achieved by the following code:
\begin{lstlisting}[language=Python]
    list_of_scenes = model.predictions_to_scenes(
        predictions=single_frame_p)
    pil_image = model.visualize_predictions(
        frames=video_frames,
        predictions=(single_frame_p, all_frame_p))
\end{lstlisting}
Note the visualization shows predictions from both heads with green and blue colors, although only ``green'' predictions from the \textit{single-frame} head are used for the evaluation.

\balance
\section{Conclusion}
This paper presents TransNet V2 -- a deep network for detection of common shot transitions, which represents an important initial step of video analysis processes. The network architecture and training experience are discussed in connection with performance evaluations comparing the network to other recent deep learning-based approaches. Simple code examples illustrate an easy usage of a selected pre-trained instance of the network. We believe that the presented software component can be easily integrated into video pre-processing pipelines of various multimedia search/analytics frameworks that require information about shots.

\begin{acks}
This paper has been supported by \grantsponsor{gacr}{Czech Science Foundation (GA\v{C}R)}{} project \grantnum{gacr}{19-22071Y} and GA UK project number 1310920.
\end{acks}

\bibliographystyle{ACM-Reference-Format}
\bibliography{references}










\end{document}